\documentclass{article}
\usepackage{ijcai22}

\usepackage{times}
\usepackage{soul}
\usepackage{url}
\usepackage[hidelinks]{hyperref}
\usepackage[utf8]{inputenc}
\usepackage[small]{caption}
\usepackage{graphicx}
\usepackage{amsmath}
\usepackage{amsthm}
\usepackage{amsfonts}
\usepackage{booktabs}
\usepackage{subcaption}
\usepackage{xcolor}
\title{Horizontal and Vertical Attention in Transformers}

\author{
Litao Yu
\and
Jian Zhang
\affiliations
University of Technology Sydney
\emails
\{litao.yu, jian.zhang\}@uts.edu.au
}

\begin{document}

\maketitle

\begin{abstract}
Transformers are built upon multi-head scaled dot-product attention and positional encoding, which aim to learn the feature representations and token dependencies. In this work, we focus on enhancing the distinctive representation by learning to augment the feature maps with the self-attention mechanism in Transformers. Specifically, we propose the horizontal attention to re-weight the multi-head output of the scaled dot-product attention before dimensionality reduction, and propose the vertical attention to adaptively re-calibrate channel-wise feature responses by explicitly modelling inter-dependencies among different channels. We demonstrate the Transformer models equipped with the two attentions have a high generalization capability across different supervised learning tasks, with a very minor additional computational cost overhead. The proposed horizontal and vertical attentions are highly modular, which can be inserted into various Transformer models to further improve the performance. Our code is available in the supplementary material.

\end{abstract}

\section{Introduction}

Transformers have achieved significant progress in various machine learning tasks \cite{NIPS17:ATTN,ICLR20:VIT,ICASSP18:SPEECH,CVPR20:MASHED}. The power of the Transformer architecture relies on learning the dependencies of tokens, which are implemented by the multi-head {\bf S}caled {\bf D}ot-{\bf P}roduct {\bf A}ttention (SDPA). In language modelling, the Vanilla Transformer model \cite{NIPS17:ATTN} learns features for each word as a token to figure out how important all other tokens are used to obtain the current features, without the use of any recurrent units. The sequential property of the word in Transformers is controlled by positional encoding, which is used in conjunction with self-attentions. These features are just weight sums and activations, so the learning process is highly parallelizable, making the Transformer model efficiently computed. When comes to the Vision Transformer (ViT) \cite{ICLR20:VIT}, we can apply a similar learning architecture by using the image patches as tokens. At the input level, the tokens are formed by uniformly splitting the image into multiple patches, e.g., splitting a $224\times224$ image into $16\times16$ patches that are $14\times14$ pixels. At intermediate levels, the outputs from the previous layer become the tokens for the next layer. At the output level, Vit applies the global average pooling followed by a multi-layer perceptron (MLP) as the classification head. ViT attains excellent performance compared to the ConvNets, and has comparably low latency in computation. However, it requires pre-training on very large-scale image datasets.

The Transformer models adopt the self-attention mechanism with the Query-Key-Value (QKV) paradigm. Given the packed matrix representations of queries $\mathbf{Q}\in\mathbb{R}^{N\times D_k}$, keys $\mathbf{K}\in\mathbb{R}^{N\times D_k}$ and values $\mathbf{V}\in\mathbb{R}^{N\times D_v}$, where $N$ is the length; $D_k$ and $D_v$ denote the dimensions of keys (or queries) and values, respectively. The self-attention is given by:
\begin{equation}\label{EQ:QKV}
\text{Attention}(\mathbf{Q},\mathbf{K},\mathbf{V})=\text{Softmax}(\frac{\mathbf{Q}\mathbf{K}^{\top}}{\sqrt{D_k}})\mathbf{V}.
\end{equation} 

The dot-products of queries and keys are divided by $D_k$ to alleviate the gradient vanishing problem of the softmax function.

A single SDPA is usually insufficient to learn distinctive features, even using the stacked Transformer blocks. Thus, the Transformer model uses multi-head attention, which is very similar to multi-view learning since each attention head focuses on a specific perspective of feature representation. Assume the input feature is $\mathbf{X}\in\mathbb{R}^{N\times D}$, where $D$ is the dimensional original queries, keys and values are projected to $D_k$, $D_k$ and $D_v$ dimensions, respectively, with $M$ parallel learned projections. For each of the projected queries, keys and values, the output is computed via Eq. (\ref{EQ:QKV}), then all outputs are concatenated and projected back to a $D$-dimensional feature representation:
\begin{align}\label{EQ:MHA}
\mathbf{Y}^M&=\text{MultiHeadAttn}(\mathbf{X}) \nonumber \\
&=\text{Concat}(\mathbf{H}_1,\ldots,\mathbf{H}_M)\mathbf{W}^M,
\end{align}
where 
\begin{align}
\mathbf{H}_m&=\text{Attention}(\mathbf{X}\mathbf{W}_m^{Q},\mathbf{X}\mathbf{W}_m^{K},\mathbf{X}\mathbf{W}_m^{V}), \nonumber\\
 \text{for}\; & m=1,\ldots, M,
\end{align}
and $\mathbf{W}^M\in\mathbb{R}^{MD_k\times D},\mathbf{W}_m^Q \in\mathbb{R}^{D\times D_k},\mathbf{W}_m^K\in\mathbb{R}^{D\times D_k},\mathbf{W}_m^V\in\mathbb{R}^{D\times D_v}$ are all projection matrices.

To build a deep model, Transformer uses a residual connection in each attention block followed by a layer-normalization:
\begin{equation}
\mathbf{Y}^O = \text{LayerNorm}(\mathbf{Y}^M+\mathbf{X}).
\end{equation}

The multi-head attention plays a key role in Transformers, for which we can simply consider it as a basic computation function just like the convolution operator. In recent works, it has been demonstrated that the model performance can be improved by explicit embedding learning functions that help capture either spatial or channel-wise correlations without requiring additional supervision \cite{TIST:ATTN}. These functions include but not limited to: Inception module \cite{AAAI17:INCEPTION}, increased cardinality \cite{CVPR17:RESNEXT}, squeeze-and-excitation \cite{CVPR18:SENET}, split-attention \cite{ARXIV:RESNEST}, spatial attention \cite{CVPR18:BUTD,ECCV18:PSA}, pyramid attention \cite{ICME20:SPA,TMM:DUAL}, etc. 

\begin{figure}[t]
\centering
\begin{minipage}{0.22\textwidth}
\centering
\includegraphics[width=1\linewidth]{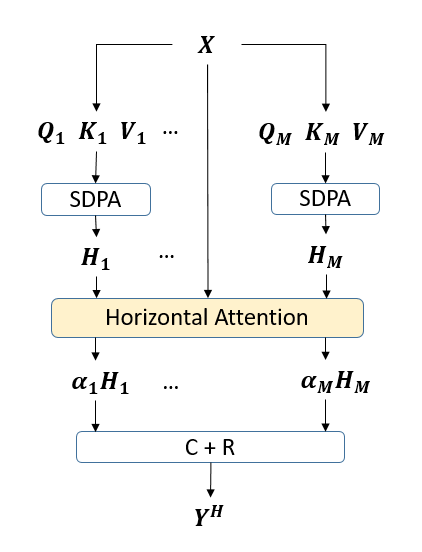}
\subcaption{Horizontal attention.}
\end{minipage}
\hfill
\begin{minipage}{0.23\textwidth}
\centering
\includegraphics[width=1\linewidth]{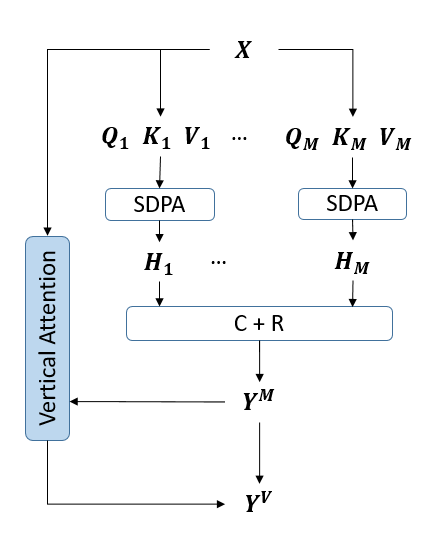}
\subcaption{Vertical attention.}
\end{minipage}
\caption{The illustration of horizontal and vertical attentions, where SDPA is Scaled Dot-Product Attention, C+R is the channel-wise concatenation then dimensionality reduction.}
\label{FIG:FRM}
\end{figure}

In this paper, we investigate two different aspects of module design in Transformers: horizontal and vertical attentions. Assume the computation of a Transformer block is a top-down pipeline, as illustrated in Figure \ref{FIG:FRM}, the top variable $\mathbf{X}$ is fed into the Transformer block, and the output is the learned feature map $\mathbf{Y}^H$ (or $\mathbf{Y}^V$) at the bottom. The horizontal attention is to re-weight the features from the multi-head attention outputs, while the vertical attention aims to re-calibrate the feature channel-wisely. The two attentions can be conducted individually or jointly, and their benefits can be accumulated through the entire Transformer learning architecture.

The development of new Transformer learning architectures is challenging engineering work, typically involving the selection of many hyper-parameters and functions. By contrast, the design of either horizontal or vertical attentions outlined above is simple and readily pluggable into existing Transformer models, whose modules can be effectively strengthened. Moreover, both horizontal and vertical attentions are light-weighted, which impose a very slight increase in model size and complexity overhead.   

Using the proposed horizontal and vertical attention in Transformers, we conducted three supervised learning tasks: machine translation, image classification and image captioning, on several public datasets. The results prove the effectiveness of the proposed methods, as well as their generalization capabilities in various Transformer models.

\section{Related work}

The Transformer model was first proposed as a sequence-to-sequence model for machine translation \cite{NIPS17:ATTN}. Later works show Transformers can also obtain very promising performance in computer vision \cite{ICLR20:VIT,ICCV21:SWIN,ICML21:DEIT} and multi-model analysis \cite{CVPR20:MASHED}. As a central piece of Transformer, the self-attention, i.e., multi-head SDPA, has the complexity and structural prior challenges. The computational complexity is mainly determined by the length of tokens, so in the long-sequence modelling, the global self-attention becomes a bottleneck for model optimization and inference. To deal with this problem, the Reformer model \cite{ICLR20:REFORMER} applies the Local Sensitive Hashing (LSH) to efficiently compute the self-attention, and Linformer \cite{ICLR20:LINFORMER} utilizes linear projection to project keys and values to a smaller length, to simultaneously reduce the complexity and model size. Inspired by the kernel approximation, Performer \cite{ICLR20:PERFORMER} follows the Random Fourier feature map to approximate Gaussian kernels. The structural prior issue mainly appears in computer vision. Unlike the invariant word embedding in natural language processing, the high uncertainty of image patches lead to the inductive bias, making Transformer models less effective than the convolution counterparts in computer vision tasks \cite{NIPS:ETVT}. To obtain a comparable accuracy with ConvNet, using Vanilla Transformer as a backbone for image classification requires the pre-training on very large-scale datasets \cite{ICLR20:VIT}. This problem can be alleviated by applying some techniques such as token distillation \cite{ICML21:DEIT}, multi-stage structures \cite{ICCV21:SWIN,ICCV21:T2TVIT} and hybrid models \cite{ARXIV:CVT}. For the improvement of multi-head attention in Transformers, Li et al. introduced an auxiliary disagreement regularization term into loss function to encourage diversity among different attention heads \cite{EMNLP18:DR}. Although some works also tried to restrict the attention spans \cite{ACL19:AAS}, or refine the aggregation \cite{NLPCC19:CAPSULE} for multi-head attention, there is no mechanism to guarantee the distinct features in Transformers. Our work aims to design extra functions to explicitly re-weight the multi-head attention and re-calibrate the feature output, to effectively improve the performance without changing the overall Transformer architectures. 

\section{Method}

In this section, we detail the two attention mechanisms that augment the feature representation of Transformers. The overall pipelines of the computation are illustrated in Figure \ref{FIG:FRM}. The proposed two methods are also based on self-attention. Considering the specific QKV paradigm with a residual connection, which is the key component in Transformer blocks, we add several feature mapping functions for the multi-head re-weighting of SDPA and channel-wise calibration.

\subsection{Horizontal attention: attention on the multi-head output of SDPA}

Multi-head attention is an appealing property in Transformers to jointly attend to the information from different feature subspaces at multiple positions. In Vanilla Transformer, the multi-head SDPA is to learn the dependencies of each token. Just like the ensemble methods \cite{WIR:ENSEMBLE} in machine learning, the multi-head attention runs SDPA several times in parallel, and each attention head is optimized independently. Their attention outputs are then concatenated and linearly transformed into an expected dimension. Intuitively, multi-head attention allows for attending to parts of the sequence differently (e.g. long-term vs short-term dependencies). The feature learning procedure naturally raises the question that which SDPA heads provide ``better'' outputs, or are relatively more important than others. It would be beneficial to focus more on these feature outputs and supress the less distinctive ones by employing a re-weighting function.

The horizontal attention is designed to auto re-weight the multi-head outputs. Specifically, we introduce a re-weighting vector $\alpha = [\alpha_1,\ldots, \alpha_M]\in\mathbb{R}^M$ that satisfies $\sum\limits_{m=1}^M \alpha_m=1$ and $\alpha_m\ge0$. Thus, the multi-head SDPA with $M$ attention heads in Eq.(\ref{EQ:MHA}) becomes:

\begin{align}
\mathbf{Y}^H &= \text{MultiHeadAttn}(\mathbf{X}) \nonumber\\
& =\text{Concat}(\alpha_1\mathbf{H}_1,\ldots,\alpha_M\mathbf{H}_M)\mathbf{W}^M.
\end{align}

To compute the re-weighting vector $\alpha$, we use the input $\mathbf{X}$ as the context variable, working with the multi-head SDPA outputs $\mathbf{H}_1,\ldots,\mathbf{H}_M$, where $\mathbf{H}_m\in\mathbb{R}^{N\times D_v}$. In simple terms, the context variable $\mathbf{X}$ acts as a dynamic representation of the relevant attention head. For each head $\mathbf{H}_m$, the horizontal attention computes a positive weight $\alpha_m$, which is interpreted as the relative importance that $\mathbf{H}_m$ provides the best output to focus for producing the feature for the subsequent computations. The pipeline to compute $\alpha$ is formulated as follows:
\begin{align}\label{EQ:ALPHA}
& \mathbf{A}_m=\text{ReLU}(\mathbf{H}_m \mathbf{W}^{A_1} + \mathbf{X}\mathbf{W}^{A_2}), \nonumber\\
& \mathbf{B}_m= \mathbf{A}_m \mathbf{W}^{B}+ \mathbf{b}^B, \nonumber\\
&\text{for}\;  m=1,\ldots, M,  \nonumber\\
& \alpha =  \text{Softmax}([\mathbf{B}_i, \ldots, \mathbf{B}_M]), 
\end{align}
where $\mathbf{W}^{A_1}\in\mathbb{R}^{D_v\times D_v}$, $\mathbf{W}^{A_2}\in\mathbb{R}^{D\times D_v}$, $\mathbf{W}^{B}\in\mathbb{R}^{D_v}$ are projection matrices, and $\mathbf{b}^B$ is a bias term.

The horizontal attention for multi-head SDPA is essentially a deterministic attention function, which corresponds to feeding in a re-weighting vector $\alpha$ to $M$ learned feature representations. Similar to the Bahdanau attention used in image captioning \cite{ICML15:SAT}, in each Transformer block, $\alpha$ is approximated by using the expected context variable $\mathbf{X}$, which can be computed by a single feedforward computation with a softmax function. This suggests that the horizontal attention approximately maximize the marginal likelihood over all multi-head SDPA outputs. Note that the computation complexity of $\alpha$ is only related to the dimension $D_v$ but irrelevant to the length of token $N$, so it can be efficiently computed even when handling the long sequences in Transformers.

\subsection{Vertical attention: attention for feature re-calibration}

The vertical attention can be viewed as a mechanism to bias the allocation of available feature channels towards the most informative components of an input signal, which is implemented by a gating function. It is specialized to model the channel-wise correlations in a computationally efficient way and designed to enhance the feature representation power of Transformer blocks throughout the whole network. Specifically, we aim to learn a channel weight vector $\beta=[\beta_1,\ldots,\beta_{D}]\in\mathbb{R}^{D}$ to re-calibrate the feature map $\mathbf{Y}^M$ in Eq.(\ref{EQ:MHA}) as:

\begin{equation}
\mathbf{Y}^V = \beta * \mathbf{Y}^M,
\end{equation}  
where $*$ is the element-wise multiplication. The computation of vertical attention is similar to the above-mentioned horizontal attention that uses the input feature representation $\mathbf{X}$ as the context variable. The difference is the vertical attention aims to re-calibrate the dimensionality reduction of the multi-head output $\mathbf{Y}^M$ in Eq. (\ref{EQ:MHA}). The computation of vertical attention is illustrated as follows:

\begin{align}\label{EQ:BETA}
& \mathbf{U}=\text{ReLU}(\mathbf{XW}^{U_1} + \mathbf{Y}^M\mathbf{W}^{U_2}), \nonumber \\
& \beta = \text{Sigmoid} (\mathbf{UW}^U + \mathbf{b}^U),
\end{align}
where $\mathbf{W}^{U_1},\mathbf{W}^{U_2}\in\mathbb{R}^{D\times D_a}$, $\mathbf{W}^{U}\in\mathbb{R}^{D_a\times D}$ are trainable mapping matrices, and $\mathbf{b}^U\in\mathbb{R}^D$ is a bias term. 

The computation of $\beta$ enables the increase of the sensitivity to the feature output $\mathbf{Y}^M$. $D_a$ is a squeezed dimension that $D_a<D$, and this setting is to embed the most informative feature component into a lower-dimensional space. Here we simply set $D_a=D/4$. The intermediate output $\mathbf{U}$ can be explained as a collection of the local descriptors whose statistics are expressive in the current context. To capture the channel-wise dependencies, it must learn a non-mutually-exclusive correlation since we need to ensure the multi-channels are allowed to be emphasized opposed to one-hot activation such as softmax activation. So in the vertical attention, we use the sigmoid activation as the gating function, which acts as the channel-weights adapted to the input-specific descriptor $\mathbf{X}$ and the learned feature representation $\mathbf{Y}^M$. The computation of vertical attention is similar to the squeeze-and-excitation (SE) module proposed in \cite{CVPR18:SENET}. The differences are: (1) In the squeeze stage of SE module, it squeezes global spatial information into a channel descriptor, which is unnecessary in the vertical attention because the feature dependencies have already been embedded by SDPA; (2) The vertical attention considers the context information given by the input $\mathbf{X}$, which jointly consider the current feature output and the context information within the Transformer block. With this regard, the vertical attention intrinsically introduces dynamics conditioned on the input and multi-head SDPA, improving the discriminative ability of feature representations.

\subsection{Complexity analysis}

To illustrate the computational efficiency of the proposed methods, we analyze the core components in both horizontal and vertical attentions in Transformers. For the multi-head SDPA in Vanilla Transformer \cite{NIPS17:ATTN}, we assume $D=D_k=D_v$, the length of the input sequence is $N$, and the number of heads is $M$. In the computation of multi-head SDPA, each head requires to store a $N\times N$ distribution matrix, so the complexity is $\mathcal{O}(MN^2 D)$. In the packed $\mathbf{Q}$, $\mathbf{K}$, and $\mathbf{V}$ of $M$ attention heads, as well as the dimensionality reduction, each Transformer block requires $2MD^2$ trainable parameters. When applying the horizontal attention on the multi-head output, the computation of re-weighting vector $\alpha$ needs extra $\mathcal{O}(MD)$ in Eq.(\ref{EQ:ALPHA}). Similarly, the computation of re-calibration vector $\beta$ in Eq.(\ref{EQ:BETA}) for vertical attention needs extra $\mathcal{O}(MD)$ to compute the intermediate variables. To store the two mapping functions to compute $\alpha$ and $\beta$, the two attentions need extra $2D^2+D$ and $3D^2$ parameters, respectively. The theoretical complexity and parameter counts are summarized in Table \ref{tab:complexity}. From the analysis, we can see that the two proposed attentions are irrelevant to the query length $N$, which means they can be readily integrated into different Transformer variants such as Performer \cite{ICLR20:PERFORMER} and Linformer \cite{ICLR20:LINFORMER}, without the change of their internal computation structures.

\begin{table}[h]
\small\centering
\begin{tabular}{lll}
\hline
 Module & Complexity & \#Params \\
\hline
Multi-head SDPA  &$\mathcal{O}(MN^2 D)$  &$2MD^2$      \\
+ Hor.  &+$\mathcal{O}(MD)$   & +$2D^2+D$      \\
+ Ver.  &+$\mathcal{O}(MD)$  & +$3D^2$     \\
\hline
\end{tabular}
\caption{Complexity and parameter counts analysis.}
\label{tab:complexity}
\end{table}

\section{Experiments}

We apply the proposed horizontal and vertical attentions in Transformer models then test them in three different learning tasks: machine translation, image classification and image captioning. We show that both attention methods can effectively boost the performance in supervised learning tasks.

\subsection{Machine translation}

\subsubsection{Datasets and experimental settings}

Machine translation is to map an input sentence representing a phrase in one language, to an output sentence representing the same phrase in a different language. In this task, we trained the Vanilla Transformer \cite{NIPS17:ATTN} on WMT-16 \cite{ARXIV:WMT16} and WMT-17 \cite{ARXIV:WMT17} English-German dataset. The relatively small WMT-16 dataset contains 29K, 1K and 1K sentence pairs for training, validation and testing, respectively. The average length of the testing target sentences is 12.4 words per sentence. The WMT-17 dataset is much larger, which has 1M, 3K and 3K sentence pairs for training, validation and testing, respectively. The testing ground truth of WMT-17 is not publicly available, so on this dataset, we report the performance on the validation split.

The attention functions were built based upon the public PyTorch implementation of Vanilla Transformer\footnote{https://github.com/jadore801120/attention-is-all-you-need-pytorch}, in which we added the horizontal and vertical attention, respectively, in the Transformer block. Note that the proposed two attention mechanisms can be either separately or jointly used in Transformer models. The word embedding dimension was set to 512 without the pre-training on extra data. The whole model is an encoder-decoder architecture, which contains 12 Transformer blocks in total. In each block, the number of SDPA heads was set to 8, and $D_k=D_v=64$. We applied the label smoothing and categorical cross-entropy. The AdamW \cite{ICLR19:ADAMW} with the default learning rate 0.001 was used to optimize the model for 100 and 120 epochs on WMT-16 and WMT-17 datasets, respectively. The experiment was conducted on a server equipped with an NVIDIA Tesla V100 GPU card. Due to the memory restriction, the mini-batch size was set to 256. 

We use the perplexity (PPL), which is the average per-word log-probability, to evaluate the machine translation quality and observe if the proposed two attention methods can bring the benefit to the Vanilla Transformer models in machine translation. The lower the PPL, the better the model is.

\subsubsection{Experimental results}

\begin{figure}[t]
\centering
\begin{minipage}{0.23\textwidth}
\centering
\includegraphics[width=1\textwidth]{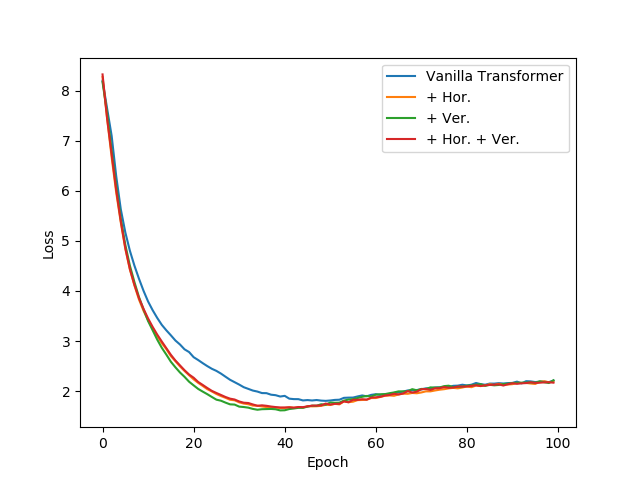}
\subcaption{Loss}
\end{minipage}
~
\begin{minipage}{0.23\textwidth}
\centering
\includegraphics[width=1\textwidth]{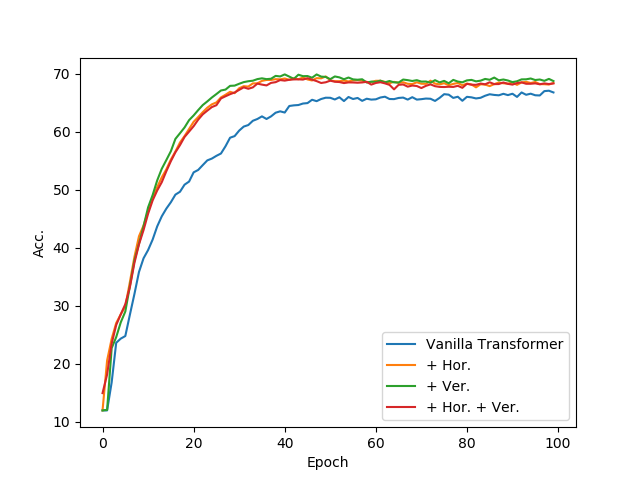}
\subcaption{Accuracy}
\end{minipage}
\caption{The loss and accuracy curves on the validation split of WMT-16 dataset.}
\label{FIG:CURVE_WMT16}
\end{figure}

\begin{figure}[t]
\begin{minipage}{0.23\textwidth}
\centering
\includegraphics[width=1\textwidth]{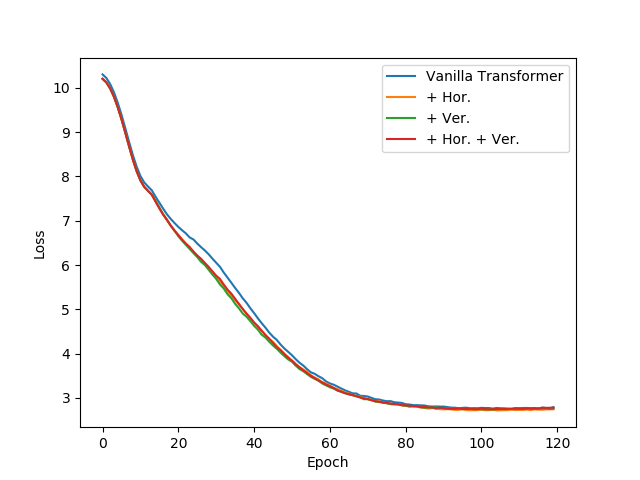}
\subcaption{Loss}
\end{minipage}
~
\begin{minipage}{0.23\textwidth}
\centering
\includegraphics[width=1\textwidth]{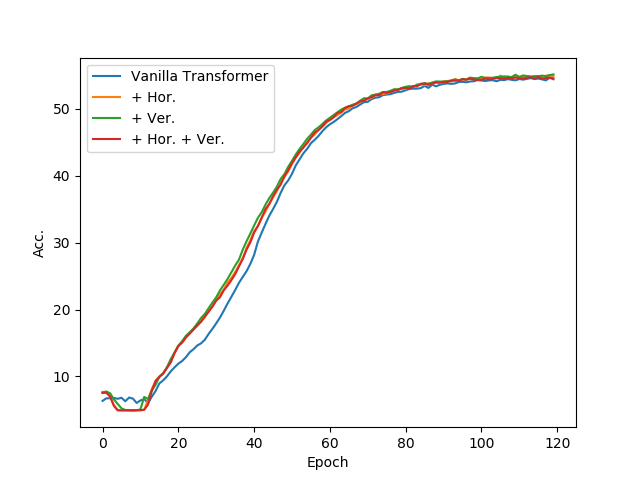}
\subcaption{Accuracy}
\end{minipage}
\caption{The loss and accuracy curves on the validation split of WMT-17 dataset.}
\label{FIG:CURVE_WMT17}
\end{figure}

\begin{table*}[t]
\small\centering
\begin{tabular}{c|ccll|ccl}
\toprule
Method  &\multicolumn{4}{c|}{WMT-16} &\multicolumn{3}{c}{WMT-17}  \\
		& \#Params &\#FLOPs  &Val PPL &Test PPL & \#Params &\#FLOPs & Val PPL \\
\midrule
Vanilla Transformer &46.7M &2.9G &6.13 &5.98 &55.8M &3.5G &16.35 \\
+ Hor. &47.1M &3.0G &5.36 \textcolor{blue}{$_{\downarrow 0.77}$} &4.99 \textcolor{blue}{$_{\downarrow 0.99}$} &56.4M &3.6G &15.53 \textcolor{blue}{$_{\downarrow 0.82}$}\\
+ Ver. &50.1M &3.1G &5.12 \textcolor{blue}{$_{\downarrow 1.01}$} &4.85 \textcolor{blue}{$_{\downarrow 1.13}$} &59.2M &3.7G &15.51 \textcolor{blue}{$_{\downarrow 0.84}$} \\
+ Hor. + Ver. &50.7M &3.2G &5.23 \textcolor{blue}{$_{\downarrow 0.90}$} &4.87 \textcolor{blue}{$_{\downarrow 1.11}$} &59.8M &3.8G &15.40 \textcolor{blue}{$_{\downarrow 0.95}$} \\
\bottomrule
\end{tabular}
\caption{Comparisons of Vanilla Transformer with its horizontal and vertical attention extensions for EN-DE machine translation on WMT-16 and WMT-17 datasets.}
\label{TAB:LANGUAGE}
\end{table*}

We plot the validation curves of different methods in Figure \ref{FIG:CURVE_WMT16} for WMT-16 and Figure \ref{FIG:CURVE_WMT17} for WMT-17 datasets, respectively. Considering the vocabulary size and number of data samples, both loss and accuracy curves in WMT-16 converge faster than in WMT-17. In terms of the validation loss and accuracy, inserting either horizontal or vertical attention, or both, can reduce the overall loss values thus improving the classification accuracy.

Table \ref{TAB:LANGUAGE} shows the model sizes, FLOPs and PPL statistics on the two machine translation datasets. Here we assume the maximal length of the input sentence is $N=64$. On the WMT-16 dataset, we can see that by adding the proposed horizontal and vertical attention functions to all Vanilla Transformer blocks, the number of trainable parameters is only increased by 0.4M and 3.4M, and the number of FLOPs is increased by 0.1G and 0.2G, respectively. Simultaneously applying the two attentions does not lead to significant computational burdens, which needs about 8.6\% and 10.3\% model parameters and FLOPs overhead. The model complexity on the WMT-17 dataset is comparably higher, due to the larger vocabulary size. However, the additional required resources of the two attentions are similar. The overall quality of machine translation is benefited by the two attention methods. On the WMT-16 dataset, horizontal and vertical attentions bring 0.99 and 1.13 PPL drops on the test split, respectively. Adding both of the two attention functions improves the baseline by 1.11 PPL, which is slightly worse than applying vertical attention. On the WMT-17 benchmark, the best validation PPL is obtained by the joint use of horizontal and vertical attentions.

\subsection{Image classification}

\subsubsection{Dataset and baseline models}

We use the ImageNet-100 \cite{ICML20:IN100}, which is a subset of ImageNet-1K, to test the effectiveness of the proposed horizontal and vertical attention in visual Transformers. Compared to ConvNets, visual Transformers lack the inductive bias and the translation invariance, so they are usually required to pre-train on large-scale image datasets \cite{ICLR20:VIT} or use in conjunction with convolutions \cite{ICML21:DEIT}. Also, in our own practice, we found that on low-resolution images such as down-sampled ImageNet (e.g., $64\times 64$) and CIFAR-100, training visual Transformers (e.g. Swin Transformer) from scratch achieves much inferior performance compared to ConvNets (e.g., ResNet38). So in our experiment, we use ImageNet-100 (100 classes) with the input size $224\times224$, where about 126K and 5K images are used for training and validation, respectively.  

Here we apply two visual Transformer models as baselines: Swin Transformer \cite{ICCV21:SWIN} and Tokens-to-token ViT \cite{ICCV21:T2TVIT}. Swin Transformer builds hierarchical feature maps by merging feature patches in down-samplings and has a linear computation complexity to feature resolution due to self-attention being only computed within local windows. In the experiment, we use Swin-T as the baseline. The Token-to-token ViT (T2T-ViT) incorporates a layer-wise Token-to-Token transformation to progressively structurize the image to tokens by recursively aggregating neighbouring tokens into one token, and an efficient backbone with a deep-narrow structure. Here we use T2T-ViT-14 with the Performer implementation as the baseline of image classification. We inserted the proposed horizontal and vertical attention modules into the Transformer blocks, providing additional functions, then observe their effectiveness. We trained the baselines, both with and without the two additional attention modules, using the same data-augmentation protocol, and applied the hyper-parameter configuration suggested by the authors of Swin-T and T2T-ViT-14. We did not tune the specific hyperparameters when we used the proposed attention functions in all the experiments. All models were trained from scratch on a single GPU card and optimized by AdamW for 200 epochs, and the mini-batch size was set to 128.

\begin{table}[t!]
\small\centering
\begin{tabular}{c|ccll}
\toprule
Method &\#Params &\#FLOPs &Acc\@@1 &Acc\@@5  \\
\midrule
Swin-T 		&26.3M &4.1G &87.52  &97.40  \\
+ Hor. 		&26.4M &4.1G &87.88 \textcolor{blue}{$_{\uparrow 0.36}$} &97.56 \textcolor{blue}{$_{\uparrow 0.16}$} \\
+ Ver. 		&27.8M &4.3G &89.76 \textcolor{blue}{$_{\uparrow 2.24}$} &98.10 \textcolor{blue}{$_{\uparrow 0.70}$} \\
+ Hor. + Ver. &28.0M &4.4G &89.06 \textcolor{blue}{$_{\uparrow 1.54}$} &97.74 \textcolor{blue}{$_{\uparrow 0.34}$}  \\
\midrule
T2T-ViT-14 	&20.1M &4.0G &86.76 &96.84  \\
+ Hor. 		&20.5M &4.2G &86.98 \textcolor{blue}{$_{\uparrow 0.22}$} &96.62 \textcolor{red}{$_{\downarrow 0.22}$} \\
+ Ver. 		&21.6M &4.3G &87.48 \textcolor{blue}{$_{\uparrow 0.72}$} &96.94 \textcolor{blue}{$_{\uparrow 0.10}$} \\
+ Hor. + Ver. &22.0M &4.4G &87.38 \textcolor{blue}{$_{\uparrow 0.62}$} &97.02 \textcolor{blue}{$_{\uparrow 0.18}$} \\
\bottomrule
\end{tabular}
\caption{Top-1 and top-5 classification accuracy and model complexity comparisons on ImageNet-100 dataset.}
\label{TAB:IMCLS}
\end{table}

\begin{table*}[t!]
\small\centering
\begin{tabular}{c|lllll}
\toprule
Method &BLEU-1 &BLEU-4 &METEOR &ROUGE-L &CIDEr  \\
\midrule
$\mathcal{M}^2$ Transformer &80.8 &39.1 &29.2 &58.6 &131.2 \\
+ Hor. &80.9 \textcolor{blue}{$_{\uparrow 0.1}$} &39.0 \textcolor{red}{$_{\downarrow 0.1}$} &29.1 \textcolor{red}{$_{\downarrow 0.1}$} &58.7 \textcolor{blue}{$_{\uparrow 0.1}$} &132.4 \textcolor{blue}{$_{\uparrow 1.2}$}\\
+ Ver. &81.0 \textcolor{blue}{$_{\uparrow 0.2}$}  &38.9 \textcolor{red}{$_{\downarrow 0.2}$} &29.3 \textcolor{blue}{$_{\uparrow 0.1}$} &58.8 \textcolor{blue}{$_{\uparrow 0.2}$} &132.9 \textcolor{blue}{$_{\uparrow 1.7}$}\\
+ Hor. + Ver. &80.9 \textcolor{blue}{$_{\uparrow 0.1}$}  &38.9 \textcolor{red}{$_{\downarrow 0.2}$} &29.3 \textcolor{blue}{$_{\uparrow 0.1}$} &58.7 \textcolor{blue}{$_{\uparrow 0.1}$} &133.2 \textcolor{blue}{$_{\uparrow 2.0}$}\\
\bottomrule
\end{tabular}
\caption{Comparisons of $\mathcal{M}^2$ Transformer with its horizontal and vertical attention extensions on the Karpathy test split of MSCOCO dataset.}
\label{TAB:IMCAP}
\end{table*}

\subsubsection{Experimental results}

\begin{figure}[t]
\centering
\begin{minipage}{0.23\textwidth}
\centering
\includegraphics[width=1\textwidth]{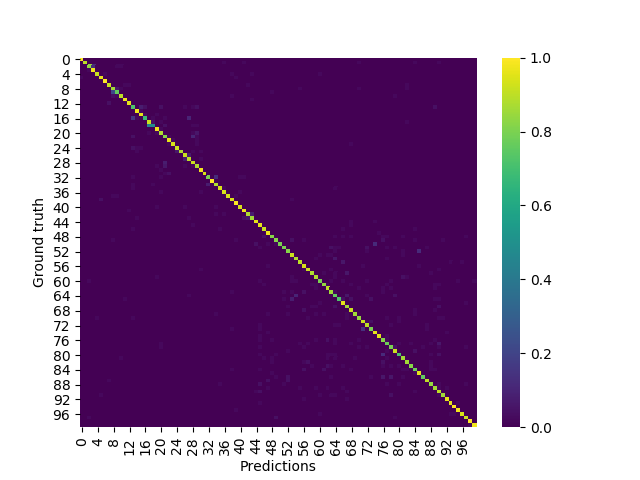}
\subcaption{Swin-T}
\end{minipage}
~
\begin{minipage}{0.23\textwidth}
\centering
\includegraphics[width=1\textwidth]{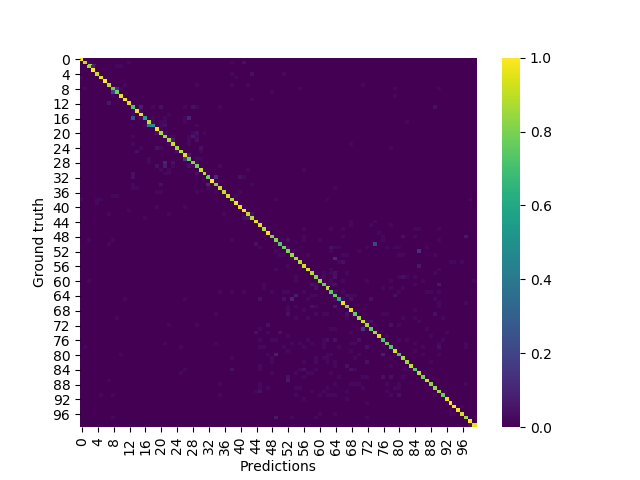}
\subcaption{+ Hor.}
\end{minipage}
~
\begin{minipage}{0.23\textwidth}
\centering
\includegraphics[width=1\textwidth]{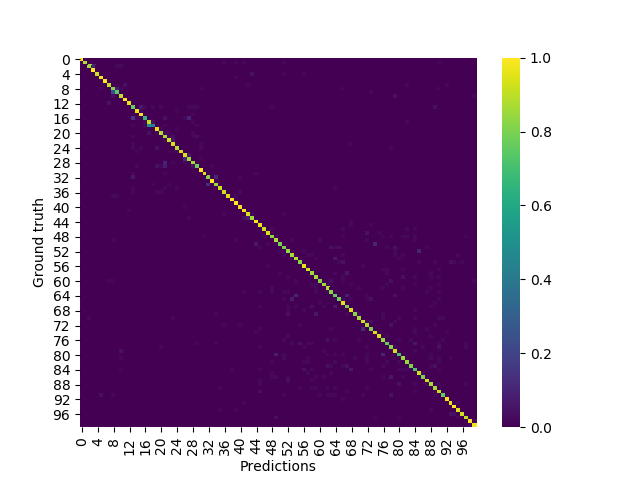}
\subcaption{+ Ver.}
\end{minipage}
~
\begin{minipage}{0.23\textwidth}
\centering
\includegraphics[width=1\textwidth]{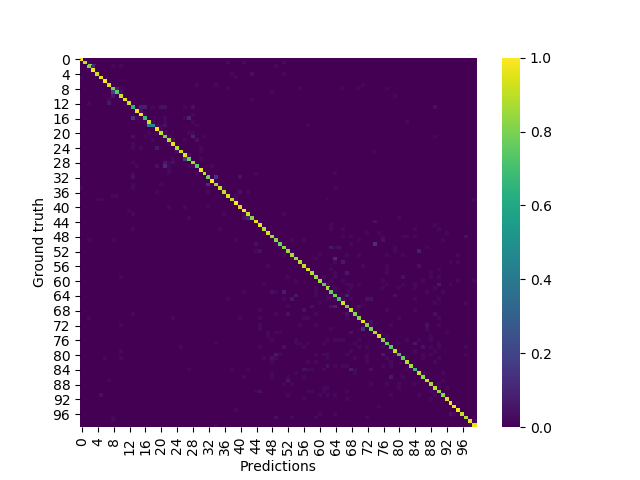}
\subcaption{+ Hor. + Ver.}
\end{minipage}
\caption{The confusion matrix of Swin-T on ImageNet-100 validation set.}
\label{FIG:CMT_SWIN}
\end{figure}

In Table \ref{TAB:IMCLS} we analyze the computational complexities and compare the classification accuracies of the two visual Transformer baselines and their extensions with horizontal or vertical attentions. In terms of model size and computational complexity, even applying both of the two proposed attentions in visual Transformers, it incurs less than 10\% additional parameters and FLOPs, which proves the high computational efficiency without changing the overall network architectures. The Swin-T generally obtains slightly higher accuracies, but T2T-ViT-14 is more parameter efficient. The classification accuracies are shown in this table, and the confusion matrices on the validation set using Swin-T are illustrated in Figure \ref{FIG:CMT_SWIN}. By observing these statistical results, we can see that the exception only occurs when applying the horizontal attention on T2T-ViT-14, where the top-5 accuracy slightly drops by 0.22\%, in all other cases, the classification accuracies are more or less improved by using the two proposed attention functions. Specifically, applying the horizontal and vertical attentions separately or jointly, the top-1 accuracy can be boosted by up to 2.24\% and 0.72\% in Swin-T and T2T-ViT-14, respectively. We also note that vertical attention, which aims to re-calibrate the output feature representations, performs better than inserting the horizontal attention into the baseline models. The independent use of vertical attention also outperforms the joint use of the two proposed attentions, in visual Transformers for image classification tasks.

\subsection{Image captioning}

\subsubsection{Dataset and model settings}

Image captioning is the task of describing the semantics of an image in natural sentences. As such, the image-to-language translation requires understanding and modelling the correlations between visual features and semantic elements, and generating a descriptive natural sentence. In this task, we tested the proposed horizontal and vertical attentions in meshed-memory ($\mathcal{M}^2$) Transformer \cite{CVPR20:MASHED} on the Karpathy split \cite{CVPR15:KARPATHY} of MSCOCO dataset. $\mathcal{M}^2$ Transformer encapsulates a multi-layer encoder and decoder, which are connected in a mesh-like structure weighted through a learnable gating mechanism via persistent memory vectors. The proposed two attention functions were built in the meshed Transformer blocks in both the encoder and the decoder. The Karpathy split of MSCOCO dataset for image captioning contains 82.8K, 5K and 5K images for training, validation and testing, respectively. Following the common practice, the models were pre-trained with cross-entropy then fine-tuned with CIDEr optimization (self-critical learning). Here we report the BLEU-1, BLEU-4, METEOR, ROUGE-L and CIDEr scores after the self-critical learning stage.

\subsubsection{Experimental results}

Table \ref{TAB:IMCAP} summarizes the performance comparisons of image captioning on the testing split. From the statistics, we can see that by applying either horizontal or vertical attentions, the BLUE-4 score actually decreases slightly, while most other evaluation scores, especially the CIDEr score, improve in accordance with the use of the two proposed attention functions. When applying the horizontal attention in Transformer blocks, the BLUE-1, ROUGE-L and CIDEr scores obtain a 0.1, 0.1 and 1.2 improvement, respectively. Inserting the vertical attention for feature re-calibration, all testing scores are improved by a certain range, except the BLUE-4 score has a further drop of 0.2. Joint integrating horizontal and vertical attentions into $\mathcal{M}^2$ Transformer obtains the highest CIDEr score, a 2.0 improvement of the baseline model. However, all other evaluation metrics are not significantly benefited from the CIDEr fine-tuning, compared to the separate use of the two proposed attention methods. The experimental results can generally prove that both horizontal and vertical attentions are able to augment the feature representation by fully exploring the visual-semantic relationships between visual features and natural sentences in Transformers.

\section{Conclusion}

We have proposed two novel self-attention methods, namely horizontal and vertical attentions, to improve feature learning in various Transformer models. The horizontal attention aims to explicitly re-weight the multi-head output of SDPA, while the vertical attention tries to re-calibrate the feature output through a gating function. The proposed methods are highly modular, which can be readily pluggable to various Transformer models for many supervised learning tasks. The experimental results show that both attention methods can effectively improve the model accuracy while requiring very limited computational resources overhead.

\bibliographystyle{named}
\bibliography{ijcai22}

\end{document}